\documentclass[11pt,a4paper]{article}

\usepackage[hyperref]{emnlp-ijcnlp-2019}
\usepackage{times}
\usepackage{latexsym}
\usepackage{graphicx}
\usepackage{xcolor}
\usepackage{amsmath}
\usepackage[utf8]{inputenc}
\usepackage{algorithm}
\usepackage[noend]{algpseudocode}
\usepackage{xparse}
\usepackage{adjustbox}
\usepackage{array}
\usepackage{booktabs}
\usepackage{multirow}

\newcommand\Tstrut{\rule{0pt}{2.0ex}}         
\newcommand\Bstrut{\rule[-0.9ex]{0pt}{0pt}}   

\makeatletter
\NewDocumentCommand{\LeftComment}{s m}{%
  \Statex \IfBooleanF{#1}{\hspace*{\ALG@thistlm}}\(\triangleright\) #2}
\makeatother
\graphicspath{ {./figs/} }

\newcolumntype{R}[2]{%
    >{\adjustbox{angle=#1,lap=\width-(#2)}\bgroup}%
    l%
    <{\egroup}%
}

\newcommand{\mpara}[1]{\medskip\noindent{\bf{#1}}}

\usepackage{url}
\usepackage{enumitem}

\newcommand{\essentia}{\mbox{\sc Essentia}}

\aclfinalcopy 

\title{Essentia: Mining Domain-Specific Paraphrases \\with Word-Alignment Graphs}

\author{Danni Ma$^{1}$, Chen Chen$^2$, Behzad Golshan$^2$, Wang-Chiew Tan$^2$\\
$^{1}$Department of Computer and Information Science, University of Pennsylvania\\
$^{2}$Megagon Labs\\
{\tt dannima@seas.upenn.edu},\quad{\tt \{chen,behzad,wangchiew\}@megagon.ai}}

\date{}

\begin{document}
\maketitle

\begin{abstract}
Paraphrases are important linguistic resources for a wide variety
of NLP applications. Many techniques for automatic paraphrase
mining from general corpora have been proposed. While these techniques are successful at
discovering generic paraphrases, they often fail to identify domain-specific
paraphrases (e.g., \{``\textit{staff}", ``\textit{concierge}"\} in the
hospitality domain). This is because current techniques are often based on
statistical methods, while domain-specific corpora are too small to fit statistical methods. In this paper, we present
an unsupervised graph-based technique to mine paraphrases from a small set of
sentences that roughly share the same topic or intent. Our system, \essentia,
relies on word-alignment techniques to create a \emph{word-alignment graph}
that merges and organizes tokens from input sentences. The resulting graph
is then used to generate candidate paraphrases. We demonstrate that our system obtains
high quality paraphrases, as evaluated by crowd workers. We further show that the
majority of the identified paraphrases are domain-specific and thus complement existing paraphrase databases.

\end{abstract}

\section{Introduction}
\label{sec:intro}
Paraphrases are important linguistic resources which are widely
used in many NLP tasks, including text-to-text generation~\cite{Juri2011}, recognizing
textual entailment~\cite{Ido2005}, and machine translation~\cite{Yuval2009}. Today, mining paraphrases still remains an active research area~\cite{ferreira2018combining, gupta2018deep, iyyer2018adversarial,
zhang2019paws}. 
Most existing work on this topic focuses on mining
general-purpose paraphrases (e.g., \{``\textit{prevalent}",
``\textit{very common}"\}),
but fails to extract \textbf{domain-specific
paraphrases}.
For example, while \{``\textit{reservation}",
``\textit{stay}''\} are not paraphrases in general, they are interchangeable in the following sentence:

\begin{center}
\small
\textit{Can we extend our reservation for two more days?}
\end{center}

\begin{figure}
    \centering
    \includegraphics[scale=0.32]{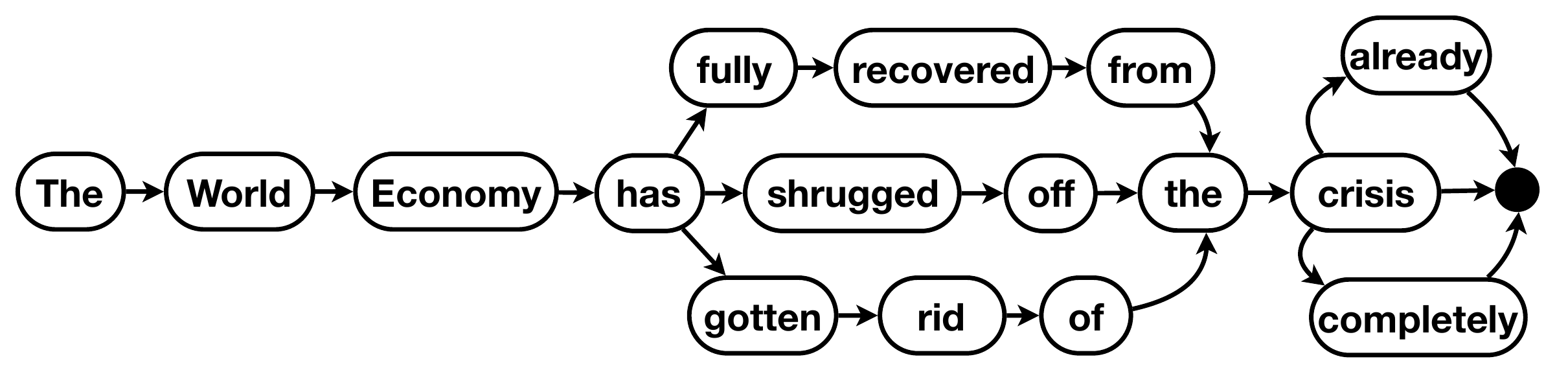}
    \caption{An instance of a word-alignment graph.}
    \label{fig:example}
\end{figure}

Existing paraphrase mining techniques are often based on statistical methods. They cannot be immediately applied to domain-specific corpora, because such corpora are usually smaller in size and lack parallel data. \essentia{} overcomes this problem by using an unsupervised graph-based method that mines domain-specific paraphrases from a small set of short sentences sharing the same topic or intent. \essentia's key insight is that a collection of sentences from a specific domain often exhibit common patterns. \essentia{} makes use of these properties to align tokens of input sentences. The resulting alignments are then summarized in a directed acyclic graph (DAG) called the \emph{word-alignment graph}. It illustrates which phrases can be used interchangeably and thus are potential paraphrases. Figure~\ref{fig:example} shows the word-alignment graph generated from the following three sentences:\\[1ex]
\begin{footnotesize}
{\em - The world economy has fully recovered from the crisis.}\\
{\em - The world economy has shrugged off the crisis completely.}\\
{\em - The world economy has gotten rid of the crisis already.}\\
\end{footnotesize}

The word-alignment graph reveals that phrases that are not 
aligned, but share the same aligned context (i.e. surrounding words) are likely to be
domain-specific paraphrases. Hence, even though \{``\textit{fully recovered from}'',
``\textit{shrugged off}'', ``\textit{gotten rid of}''\} are not aligned, they are likely paraphrases because they
share the same patterns before and
after themselves. 

While this work is focused on mining paraphrases, we believe that word-alignment graphs
have other interesting applications, and we leave them for future work. For instance, a
word-alignment graph enables one to generate new sentences or phrases that do
not appear in the original set of sentences.
``\textit{The world economy has gotten rid of the crisis completely}''
is a new sentence that is generated using the graph in
Figure~\ref{fig:example}.

\mpara{Contributions.}\quad
We present \essentia, an unsupervised system for mining domain-specific paraphrases by creating rich graph structures from small corpora.
Experiments on datasets in real-world applications demonstrate
that \essentia{} finds high-quality domain-specific paraphrases. We also validate that these domain-specific paraphrases complement and augment PPDB (Paraphrase
Database), the most extensive paraphrase database available in the community.

\section{Essentia}
\begin{figure}
    \centering
    \includegraphics[scale=0.28]{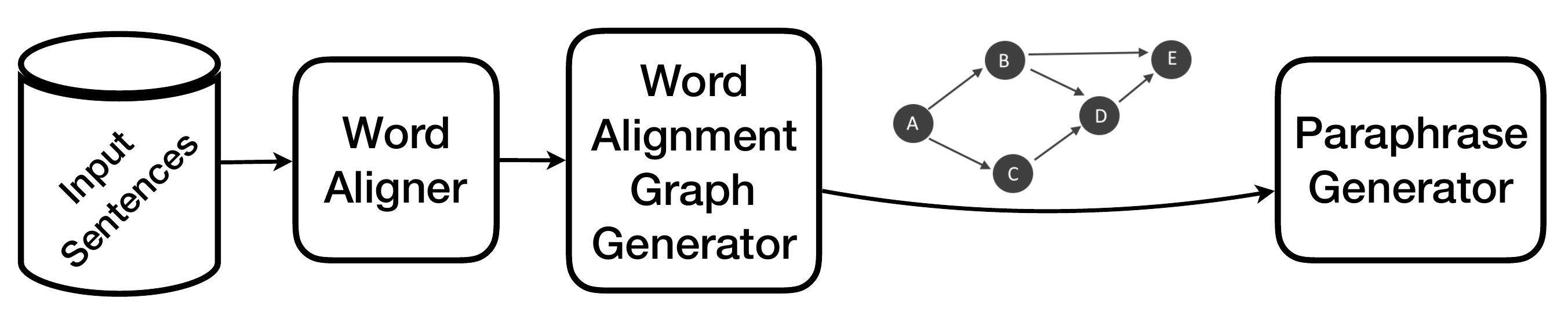}
    \caption{The architecture of \essentia.}
    \label{fig:arch}
\end{figure}

The architecture of \essentia{} (Figure~\ref{fig:arch}) consists of:
(1) a word aligner which aligns similar words (and phrases) 
between different sentences based on syntactic and semantic similarity;
(2) a word-alignment graph generator that summarizes the alignments into a compact
graph structure; and (3) a paraphrase generator that mines
domain-specific paraphrases from the word-alignment graph. We describe each component below.

\subsection{Word aligner}
We use the state-of-the-art monolingual 
word aligner by
\citet{sultan2014back}. The input to the word aligner is a single pair of
sentences and the output is a predicted mapping between tokens of
two sentences. \essentia{} uses the word aligner to compute the alignments for
all pairs of sentences provided as input. 


Every sentence is first pre-processed by 
replacing numbers and named entities -- which are identified by spaCy \cite{honnibal2017spacy} -- with special symbols ``NUM” and 
``ORG” respectively before it is passed to the word aligner. 

The word aligner relies on paraphrase, lexical resources and word embedding techniques to find a mapping
between tokens. In other words, the word aligner finds general-purpose paraphrases and maps their tokens accordingly.
\essentia{} further processes the output of the word aligner to mine domain-specific paraphrases.

\subsection{Word-alignment graph generator}
\label{arch:graphgen}
Once the alignments between every pair of sentences are available, the 
word-alignment graph generator summarizes all the alignments into a unified structure, referred to as the word-alignment graph. It is a 
DAG that represents all the input sentences (see Figure~\ref{fig:example} as
an example). The process of creating the word-alignment graph is described as follows.

The first step partitions the set of input sentences into
\emph{compatible} groups. A group of sentences is compatible if
their alignments adhere to the following three conditions:
\begin{itemize}[leftmargin=*]
    \itemsep0em 
    \item \textbf{Injectivity}\quad For any pair of sentences, each word should be
    mapped to at most one word in the other sentence.
    \item \textbf{Monotonicity}\quad For any pair of sentences, if a word $w1$ appears
    before $w2$, then the word that $w1$ maps to should also appears before the
    word that $w2$ maps to in the other sentence. Sentence pairs such as 
    ``{\em Yesterday I saw him}" and ``{\em I saw him yesterday}" 
    violate this condition.
    \item \textbf{Transitivity}\quad Given any three sentences $s_1$, $s_2$, and $s_3$,
    if a word $w_1$ in $s_1$ is mapped to $w_2$ in $s_2$, and $w_2$ in $s_2$ is mapped to
    $w_3$ in $s_3$, then $w_1$ should be only mapped to $w_3$ in $s_3$.
\end{itemize}

The above conditions are necessary to ensure that the resulting representation is compact and forms a DAG.
We start by partitioning the input sentences into compatible groups. 
The partitioning strategy is a simple greedy algorithm which starts with a single empty group.
A sentence will be added to the first group that remains compatible upon adding this new sentence.
If no such group exists, a new empty group is created and the sentence is added
to this group. 
This process repeats until each sentence is assigned to one group.

Next, the word-alignment graph generator represents each
group as a DAG and then combines all the DAGs using a shared start-node and 
end-node to create the final word-alignment graph. 
Specifically, a line graph is first created for each sentence
(i.e., a word-alignment graph for a single sentence). Then,
the alignments are processed: for each pair of aligned words, their corresponding nodes are contracted to a single node. Due to the constraints imposed
earlier, one can easily show that the resulting graph will be cycle-free.


\subsection{Paraphrase generator}
Given a word-alignment graph, the paraphrase generator considers all paths in the graph that share the same start and end node as paraphrase candidates. For
instance, in Figure~\ref{fig:example}, there are three branches that start from 
the node ``\textit{has}" and end in ``\textit{the}". Consequently, the phrases \{``\textit{fully recovered from}'', ``\textit{shrugged off}'', 
``\textit{gotten rid of}''\} are extracted as paraphrase candidates. 

However, not all extracted candidates are paraphrases. Consider the following sentences: \\[1mm]
\begin{small}
{\em - Give me directions to my parent's place}\\
{\em - Give me directions to the Time Square}
\end{small}

\noindent 
In this case, \{``\textit{my parent's place}'',
``\textit{the Time Square}''\} will be extracted as candidates, but it is
clear that they are not valid paraphrases.

To avoid generating wrong paraphrases, we design a filtering step -- which can be implemented either using rules (e.g., regular expressions) or statistical methods (e.g., word similarity) -- on top
of the extracted candidates. 
Our current implementation of this filtering functionality adopts a rule-based heuristic that only considers candidates of verb phrases containing three or fewer tokens, such as \{``\textit{access to Wi-Fi}'', ``\textit{hookup to Wi-Fi}''\}. Our empirical study reveals that many such verb phrases are domain-specific paraphrases. Other classes of phrases, such as noun phrases, turn out to have much noise. For example, many noun phrases are simply different options (e.g., \{``\textit{today}'',``\textit{tomorrow}''\}). We leave the design of advanced filters for those classes as future work. 

In the process of discovering paraphrases, we observe that sentences can be ``cleaned''. That is, 
some phrases can be removed without affecting the essential meaning of a sentence. 
Figure~\ref{fig:example} shows that the phrases ``\textit{already}" and ``\textit{completely}" share the same start and end node. Moreover, we see that
the start and end node are also directly connected with a single edge. Such phrases are {\em optional phrases} and can be removed without affecting the core meaning of a sentence.
By identifying optional phrases, we can
simplify the set of input sentences to its ``essence'', where the name of \essentia{} comes from.

\mpara{Notes on scalability.}\quad
The time required by the word aligner to compute alignments between two sentences is quite small and can be considered as constant since the length of input sentences is bounded in practice. Given that, the
time-complexity of \essentia's pipeline for $n$ input sentences is $O(n^2)$ as we need to compute alignments between all pairs of sentences.
In practice, the pipeline can be applied to roughly
a hundred sentences within an hour. For a larger collection of sentences, as described in Section~\ref{arch:graphgen}, we first run a clustering algorithm to group sentences into smaller clusters, and then feed each cluster to \essentia's pipeline.

\section{Related Work}
\begin{table*}[t]
\small
    \centering
    \captionsetup{justification=centering}
    \begin{tabular}{c|c|c|c|c}
    \hline
    \multicolumn{1}{c}{\textbf{}} & \textbf{Dataset} & \textbf{\# of extracted pairs} & \textbf{\# of valid pairs} & \textbf{Precision} \\ \hline \hline
    \multirow{2}*{\essentia{}} & Snips & 173 & 84 & 48.55\% \Tstrut \\
    \cline{2-5}
    ~ & HotelQA & 2221 & 642 & 28.91\% \Tstrut \\ \hline
    \multirow{2}*{FSA} & Snips & 18 & 15 & 83.33\% \Tstrut \\
    \cline{2-5}
    ~ & HotelQA & 342 & 185 & 54.09 \% \Tstrut \\ \hline
    \end{tabular}
    \caption{Comparison between \essentia{} and FSA baseline on paraphrase extraction}
    \label{tab:results}
\end{table*}

Collecting and curating a database of paraphrases is a
costly and time-consuming task in general. Although there
are existing techniques to collect paraphrase pairs
from crowd-workers more efficiently and with lower 
cost~\cite{chen11collecting}, there has been a great
interest in developing 
techniques for automatically mining paraphrases from existing
corpora. \citet{barzilay2001extracting} proposed the first
unsupervised learning algorithm for paraphrase acquisition from a corpus of
multiple English translations of the same source text. \citet{barzilay2003learning} followed up with an approach that applied multiple-sequence alignment
to sentences gathered from parallel corpora. \citet{pang2003syntax}
proposed a new syntax-based algorithm to produce word-alignment graphs for sentences.
Finally, \citet{quirk2004monolingual} applied statistical machine translation
techniques to extract paraphrases from  monolingual parallel corpora.

The most extensive resource for paraphrases today is PPDB~\cite{ganitkevitch2013ppdb, ganitkevitch2014multilingual,
pavlick2015ppdb}. PPDB consists of a huge number of phrase pairs with confidence
estimates, and has already been proven effective for multiple tasks. However, as our experiments show, PPDB and other resources fail to capture a large number of domain-specific paraphrases.


To extract domain-specific paraphrases,
\citet{pavlick2015domain} extended Moore-Lewis method \cite{moore2010intelligent} and learned paraphrases
from bilingual corpora. \citet{ zhang2016extract} constructed Markov networks of
words and picked paraphrases based on the frequency of co-occurrences. However, these
systems rely on significantly large amounts of domain-specific data (either for supervised
training or conducting frequency analysis), which may not always be available. \essentia{} instead uses an unsupervised graph-based technique for paraphrase mining and does not rely on the presence of a large amount of domain-specific data.
The word-alignment graph constructed by \essentia{} can be interpreted as an
extension of multi-sentence compression \cite{filippova2010multi}. We compactly
maintain all paths and expressions in the constructed word-alignment graphs.
As pointed out in \citet{pang2003syntax}, the extracted paraphrases can help enrich the diversity of expressions regarding a specific intention, and ultimately provide more training examples for data-driven models.



\section{Evaluation}
\essentia{} is evaluated on two datasets and is shown to  generate
high quality domain-specific paraphrases. We compare our system against a syntax-based alignment technique by \citet{pang2003syntax}, which we refer to as FSA, as it generates \emph{Finite-State Automata} for compactly representing sentences in a setting similar to ours. Compared to FSA, \essentia{} generates 263\% more paraphrases on those two datasets. We further demonstrate that most extracted paraphrases are truly domain-specific and thus are missing from PPDB.

\smallskip
\noindent
{\bf Datasets}
We use two datasets to evaluate \essentia. The first one, commonly known
as the Snips dataset \cite{coucke2018snips}, is a collection of queries
submitted to smart conversational devices (e.g., Google Home or Alexa). Snips has
ten documents, each covering one intent such as ``\textit{Get Directions}'', ``\textit{Get Weather}'' and so on. On average, each document has 32 sentences, and each sentence has 9 words. 
The other dataset -- which is called HotelQA -- is an industry proprietary dataset of various types of questions 
submitted by hotel guests regarding different amenities and services, such as ``\textit{Check-out}'' or ``\textit{Wi-Fi}''. HotelQA also consists of ten documents, with an average of 54 sentences per document and 10 words per sentence.
HotelQA was our primary motivation for investigating this problem. The industry application requires an automatic method to identify a set of questions that are semantically equivalent.

\subsection{Mining Paraphrases}
Table~\ref{tab:results} compares the performance of \essentia{} with the FSA baseline for paraphrase mining. 
Specifically, we show the number of phrase pairs extracted by \essentia{} and FSA from both datasets (``\# of extracted pairs'' column), number of valid paraphrases within these pairs (``\# of valid pairs'' column), and precision (``Precision'' column).
Although FSA has higher precision due to conservative sentence alignment, \essentia{} extracts significantly more paraphrases, improving the recall by 460\% (Snips) and 247\% (HotelQA) over the baseline.
To identify valid paraphrases, we design a crowd-sourcing task on Figure-Eight Data Annotation Platform. In this task, we present an extracted candidate pair (e.g., \{\textit{``log onto''}, \textit{``connect to''}\}) and a domain (e.g., ``Wi-Fi'') to human annotators, and ask them to decide whether the two phrases are paraphrases or not.

\essentia{} discovers a large number of paraphrases missing from PPDB, which has the highest coverage among the existing paraphrase
resources~\cite{pavlick2016simple}. More precisely, we take the
726 correct extractions of \essentia{} (as verified by human annotators) and
search to see if they appear in PPDB even with low confidence scores.
We find that only \textbf{4\%} of our discovered paraphrases appear in PPDB. This in 
turn shows the effectiveness of \essentia{} in discovering paraphrases, because it goes beyond PPDB by
using only a few sentences.
Table~\ref{tab:para_exp} lists some domains and examples of domain-specific
paraphrases detected by \essentia. 

Finally, to better understand how \essentia's performance can be improved and what opportunities lie ahead for further research, we review a sample of \essentia{}'s incorrect extractions and identify two major classes of errors. One class consists of expressions that are alternative options but not necessarily
paraphrases (e.g., \{\textit{``avoiding the highway''},
\textit{``avoiding toll road''}\}). Another class contains expressions that
involve the same topic but have slightly different intentions
(e.g., \{``\textit{tell me the Wi-Fi password}'',
``\textit{how to connect to Wi-Fi}"\}). While the two error classes we discuss here are the most prevalent ones, an in-depth analysis of error classes and their frequencies (which we leave as future work) can be quite insightful.



\begin{table}[t]
\small
    \centering
    \begin{tabular}{||l|l||}
    \hline
    \textbf{Domain}  &  \textbf{Example paraphrases} \Bstrut \Tstrut\\ \hline \hline
    Restaurant search  &  recommend a good place \Tstrut \\ 
    & suggest a place \Bstrut\\ \hline
    Restaurant reservation  &  get me a place  \Tstrut\\ 
      & get me a spot  \Bstrut\\ \hline
    Get directions   &  show me the way \Tstrut\\ 
    & get me directions \Bstrut\\ \hline
    Get weather   &  need the weather \Tstrut\\ 
    & want the weather \Bstrut\\ \hline
    Request ride   &  find a taxi \Tstrut\\ 
    & need an uber \Bstrut\\ \hline
    Share location   &  share my location \Tstrut\\ 
    & send my location \Bstrut\\ \hline
    Hotel Wi-Fi   &  log onto the Wi-Fi \Tstrut\\ 
    & connect to Wi-Fi \Bstrut\\ \hline
    Hotel checkout   &  extend our checkout \Tstrut\\ 
    & have a late checkout \Bstrut\\ \hline
    \end{tabular}
    \caption{Examples of domain-specific paraphrases.}
    \label{tab:para_exp}
\end{table}



\section{Conclusion and Future Work}

We present \essentia{}, an unsupervised graph-based system for extracting domain-specific
paraphrases, and demonstrate its effectiveness using datasets in real-world applications. Empirical results show that \essentia{} can generate high quality domain-specific paraphrases that are largely absent from mainstream paraphrase databases.

Future work involves various directions. 
One direction is to derive domain-specific sentence templates from corpora. 
These templates can be useful for natural language generation in question-answering systems or dialogue systems.
Second, the current method can be extended to mine paraphrases from a wide range of syntactic units other than verb phrases.
Also, the word aligner can be improved to align prepositions more accurately, so that the generated alignment graph would reveal more paraphrases.
Finally, \essentia{} can also be used to identify linguistic patterns other than paraphrases, such as phatic expressions (e.g., ``\textit{Excuse me}'', ``\textit{All right}''), which will in turn allow us to identify the essential constituents of a sentence.





\bibliography{main}
\bibliographystyle{acl_natbib}

\end{document}